# Prompting Strategies for Language Model-Based Item Generation in K–12 Education: Bridging the Gap Between Small and Large Language Models

Mohammad Amini[1*], Babak Ahmadi[1*], Xiaomeng Xiong[2], Yilin Zhang[3], Christopher Qiao[4]

[1]Industrial and Systems Engineering Department, [2]College of Education,
[3]Department of Mechanical & Aerospace Engineering,
[4]Department of Computer & Information Science & Engineering,
University of Florida, Gainesville, Florida, US.

**Abstract**

This study explores automatic generation (AIG) using language models to create multiple choice questions (MCQs) for the morphological assessment, aiming to reduce the cost and inconsistency of manual test development. The study used a two-fold approach. First, we compared a fine-tuned medium model (Gemma, 2B) with a larger untuned one (GPT-3.5, 175B). Second, we evaluated seven structured prompting strategies, including zero-shot, few-shot, chain-of-thought, role-based, sequential, and combinations. Generated items were assessed using automated metrics and expert scoring across five dimensions. We also used GPT-4.1, trained on expert-rated samples, to simulate human scoring at scale. Results show that structured prompting, especially strategies combining chain-of-thought and sequential design, significantly improved Gemma's outputs. Gemma generally produced more construct-aligned and instructionally appropriate items than GPT-3.5's zero-shot responses, with prompt design playing a key role in mid-size model performance. This study demonstrates that structured prompting and efficient fine-tuning can enhance midsized models for AIG under limited data conditions. We highlight the value of combining automated metrics, expert judgment, and large-model simulation to ensure alignment with assessment goals. The proposed workflow offers a practical and scalable way to develop and validate language assessment items K-12.

## 1 Introduction

Generating high-quality test items is essential for diverse applications in educational measurement, encompassing both large-scale assessments and instructional uses (Attali, 2018). However, manual item writing is time-consuming and resource-intensive. It requires professional expertise, iterative revisions, and extensive piloting, which make large-scale or adaptive test development difficult to sustain (Bandalos, 2018; Kurdi et al., 2020).

Automated Item Generation (AIG) offers a scalable alternative. With the advent of large language models (LLMs), generating fluent, well-structured items has become more feasible (Drasgow et al., 2006; Pugh et al., 2016). Yet, key limitations remain: LLMs often fail to align with target constructs, produce unreliable distractors, and lack mechanisms for difficulty control.

For example, when prompted with *"Generate a multiple-choice question targeting the prefix mis-"*, GPT-3.5 produced:
*Which word uses the prefix mis- to show 'doing something the wrong way'?*
A. *mismatch*, B. **misuse** (correct), C. *misconduct*

While structurally sound, this item may test lexical familiarity with *match*, *use*, and *conduct*, rather than morphological awareness of the prefix *mis-*.

Moreover, vague prompts led to inconsistent results. For example, the model produced items that test whether *mis-* is a real prefix (e.g., *mission*), probe its meaning, or ask students to infer its function in context. This kind of construct confusion undermines item reliability.

Distractor quality is also a concern. In some cases, options like *misconduct* may seem more appropriate than the intended answer, potentially misleading high-performing students. In others, distractors differ morphologically (e.g., *dis-* vs. *mis-*) and become too easy to rule out, or are so obscure they're dismissed due to unfamiliarity. These issues undermine item validity and discrimination.

A further challenge is item difficulty control. LLMs lack reliable ways to generate tiered items, limiting their ability to differentiate students.

Therefore, to realize the potential of AIG in K-12 morphological assessment, it is essential to ensure construct consistency, improve distractor quality, and develop methods for fine-grained control over item difficulty. This study focuses on addressing these core challenges.

---

*[*]Mohammad Amini and Babak Ahmadi contributed equally to this work and are co-first authors.

We study automated multiple choice questions (MCQs) generation for morphological assessment using 268 WordChomp items, comparing a fine-tuned mid-sized model (Gemma, 2B) and a large instruction-following model (GPT-3.5, 175B) across seven prompting strategies: zero-shot, few-shot, chain-of-thought (CoT), role-conditioned, sequential, and hybrid.

Item quality is evaluated using both automated metrics (grammar, complexity, readability, fluency) and expert ratings across five dimensions: instruction clarity, answer accuracy, distractor quality, word difficulty, and task difficulty. We also employ GPT-4.1, trained on expert-labeled data, to simulate large-scale human-aligned scoring.

Results show that structured prompting notably enhances item quality for Gemma. While GPT-3.5 excels in surface-level fluency, Gemma yields items better aligned with morphological constructs and grade-level appropriateness. These findings highlight the potential of mid-sized models, when supported by effective prompting and efficient fine-tuning, to generate valid and scalable assessment items for K–12 contexts[1].

## 2 Related Work

Despite the progress in AIG, three core challenges persist: (1) maintaining construct validity in generated items, (2) leveraging small language models(SLMs) under limited data conditions, and (3) applying and evaluating structured prompting in linguistically complex domains. We review prior work addressing each of these.

### 2.1 Limited Construct Validity in LLM-Generated Items

Large language models such as GPT-3.5, GPT-4, PaLM, and Claude generate fluent, high-quality K–12 assessment items, producing questions that educators find suitable for classroom use (Zhuge et al., 2025). They often outperform earlier small-scale models in linguistic coherence and reasoning (Anonymous ACL submission, 2025). However, these models can misalign with curricular objectives and struggle to target specific cognitive levels. They may also introduce unintended vocabulary or extraneous clues if prompts are not carefully constrained. Distractor quality is another concern: incorrect options generated by LLMs often fail to mimic common student misconceptions and can even yield multiple plausible answers without proper validation (Scarlatos et al., 2024).

To address these limitations, researchers have developed strategies such as few-shot exemplars and chain-of-thought prompting to guide model outputs more precisely (Wang et al., 2025). Iterative refinement workflows, where a model critiques and improves its draft questions, further enhance clarity and alignment. Dual-LLM frameworks like TwinStar—pairing question generation with automated evaluation—have shown significant gains in adhering to target cognitive levels and knowledge points, even when using smaller models.Despite these advances, LLMs remain prone to hallucinations and are highly sensitive to prompt formatting, with minor changes causing substantial shifts in output quality (Sclar et al., 2024). Additionally, the computational cost of running large models at scale remains a significant challenge (Mucciaccia et al., 2025).Beyond assessment item generation, transformer-based models have also shown promise in analyzing educational discourse more broadly, with frameworks like PEER HELPER demonstrating how RoBERTa can systematically categorize mentor-mentee interactions to improve educational support systems (Carroll et al., 2025). Similarly, LLM-enhanced hybrid approaches have been applied to develop intelligent recommendation systems for personalized learning experiences, such as matching students with appropriate work-integrated learning opportunities (Hwang et al., 2025).

### 2.2 SLMs in Low-Resource AIG

Although LLMs have demonstrated outstanding capabilities across various tasks, their high computational and training costs hinder its widespread adoption (Gunasekar et al., 2023; Hu et al., 2024). In response, researchers have begun exploring SLMs as a more affordable and scalable alternative.

For instance, Syromiatnikov et al. (2025) combined parameter-efficient fine-tuning (PEFT) with CoT prompting, applying quantization and LoRA techniques to LLaMA (3B/8B) and Gemma (9B). Their models outperformed GPT-4o mini and Mistral Large on syntactic and reasoning tasks within the Ukrainian ZNO-Eval benchmark.

To improve training efficiency, Li et al. (2021) introduced a curriculum learning approach based on input length. This strategy enhanced the stability and efficiency of GPT-2 models (117M and 1.5B) under large-batch, high-learning-rate conditions, reducing training time by up to 70%.

---

[1]All code and resources are available at our GitHub repository: https://github.com/mhmdamini/NLP

Nonetheless, SLMs still face notable challenges in generation tasks. For example, Rodriguez-Torrealba et al. (2022) developed an end-to-end MCQ generation pipeline using T5-small. Expert evaluations found the items leaned toward recall-based cognition and lacked higher-order reasoning, suggesting limitations in content depth control.

Similarly, the Phi-1 model (Gunasekar et al., 2023) outperformed larger models but struggled with stylistic diversity, robustness in natural language, and multilingual generalization. These findings point to persistent constraints related to training data quality and scope.

Building on these insights, our study explores how structured prompting strategies enhance small models' generative capabilities in low-resource settings. We aim to reduce reliance on large-scale fine-tuning while assessing whether SLMs can match or exceed zero-shot LLM performance.

## 2.3 Structured Prompting Strategies

Recently, prompt-based learning has emerged as a prominent approach for bridging pretrained language models with downstream tasks. Compared to the traditional pretraining and fine-tuning framework, it reframes tasks as text completion using natural language templates, enabling strong zero- and few-shot performance (Liu et al., 2023). However, simple prompts often fall short for tasks requiring multi-step reasoning or structured outputs.

To address this, researchers have developed structured prompting strategies to improve model reasoning. CoT prompting explicitly guides models to produce intermediate reasoning steps, enhancing performance on complex tasks (Wei et al., 2022). Tree-of-Thoughts (ToT) (Yao et al., 2023) builds on CoT by structuring reasoning as a tree of candidate steps, using breadth-first or depth-first search strategies to identify optimal paths.

Other work has explored prompt framing. Role-Play Prompting assigns identities (e.g., teacher or expert), implicitly enhancing task understanding. (Kong et al., 2023), especially when reasoning chains don't naturally arise. Another category of strategies emphasizes process control. Iterative Self-Refinement allows models to draft, critique, and revise their own outputs based on natural language feedback, improving coherence without external supervision (Madaan et al., 2023).

These structured prompting strategies expand the reasoning capabilities of language models from multiple angles. However, most existing research has focused on individual methods without systematic comparison. Furthermore, structured prompting has primarily been applied to mathematical and logical tasks, with limited investigation in linguistically complex applications such as morphological analysis and reading comprehension.

This study addresses these gaps by systematically comparing structured prompting strategies in low-resource settings and evaluating how well they support generating morphological MCQs. Specifically, we examine how these methods support construct alignment, improve distractor quality, and enable fine-grained control over item difficulty.

## 3 Research Motivations

Building on the gaps identified in Section 2—most notably the limited adaptation of language models to K–12 linguistic requirements and the insufficient emphasis on construct validity—we articulate the following research motivations and guiding questions: First, can language models be leveraged to generate high-quality, developmentally appropriate MCQs that align with morphological constructs in a K–12 context? Second, how do different prompting strategies (ranging from zero-shot to structured chain-of-thought) affect the reliability and pedagogical quality of the generated items? And finally, can a fine-tuned, small-scale language model, when guided by carefully designed prompts, approximate or match the performance of larger language models, thereby reducing computational overhead without sacrificing item quality?

These questions drive our methodological choices, including prompt design, model selection, and evaluation metrics. By systematically examining each question, we seek both theoretical insights on the capabilities of LLM-based AIG and practical guidelines for implementing such systems in K–12 morphological assessments.

## 4 Dataset

The dataset comprises 268 MCQs from the *Word-Chomp* diagnostic assessment, targeting morphological skills for students in grades 3–5. Such skills are assessed by tasks focusing on identifying word parts (recognition), interpreting their meaning or function (comprehension), or applying them to new word problems (problem-solving) (Carlisle, 1995; Apel and Henbest, 2016). Examples appear in Appendix A and B.

Each item has a stem averaging 8.76 words (minimum 5, maximum 16), spanning *comprehension* (40 items), *problem-solving* (108), and *recognition* (120). Morphologically, 173 items are *derivational*, 47 *inflectional & derivational*, 23 *inflectional*, 19 *define*, 4 *syntactic*, and 2 address *word parts*.

Word difficulty was originally rated on a 1–5 scale in 0.5 increments. For analysis, half-points were merged with the lower integer (e.g., 1.5 became 1), yielding five levels: *Level 1* (37 items), *Level 2* (63), *Level 3* (87), *Level 4* (69), and *Level 5* (12). Task difficulty was recoded similarly, resulting in *Easy* (99), *Medium* (103), and *Hard* (66) items. A Spearman correlation ($\rho = 0.469$, $p < 0.001$) indicated a moderate positive relationship between word and task difficulty.

Items cover varied morphological challenges, such as choosing a correct prefix, suffix, or root (81 items), breaking words into parts (39), and checking spelling or meaning based on morphemes (e.g., inflected vs. derived forms). The dataset comprises 13 question types (QT1–QT13), detailed in Appendix A, allowing comprehensive analysis of how language models handle varying morphological complexity, from straightforward recognition to more intricate affix-based transformations.

## 5 Methodology

We adopt a two-fold approach that (1) leverages multiple language models of varying sizes, and (2) employs structured prompting techniques to elicit high-quality morphological MCQs.

### 5.1 Fold One: Language Models

In this first fold, we investigate both *small language models* (SLMs) and a *large language model* (LLM) to systematically compare how parameter scale and pre-training scope influence morphological item generation. While large models like GPT-3.5 are powerful general-purpose tools, recent studies have demonstrated that smaller, fine-tuned models can outperform larger models in domain-specific tasks. For instance, (Bucher and Martini, 2024) showed that fine-tuned smaller LLMs consistently and significantly outperform larger, zero-shot prompted models in text classification tasks across diverse categories, including sentiment and emotion analysis. Similarly, (Hsieh et al., 2023) introduced a distillation method that enables smaller models to surpass larger ones with less training data. These findings suggest that carefully fine-tuned mid-scale models can approximate or even exceed the performance of larger LLMs in specialized domains, offering benefits in computational cost, data efficiency, and output fidelity.

#### 5.1.1 Small Language Models (SLMs).

We evaluate three smaller-scale language models for morphological question generation: T5 (60M), a bidirectional encoder-decoder suited for understanding contextual relationships; GPT-2 (117M), a unidirectional autoregressive model effective for text generation; and Gemma (2B), a medium-scale decoder with advanced attention mechanisms, allowing us to explore the impact of model size.

**Model Fine-tuning Process.** To comprehensively evaluate language models of different scales and architectures on the morphological assessment item generation task, we implemented specialized fine-tuning and evaluation strategies for three distinct models. For all models, we implemented consistent data preprocessing steps with specific adaptations for each model's characteristics. The 268 WordChomp items were first converted into formats compatible with each model architecture. Next, we implemented an 80-10-10 stratified split (training-validation-testing) to ensure proportional representation of all 13 question types and difficulty levels. For T5, inputs were represented as a "question generation task" with the correct answer as output. For GPT-2, complete question-answer pairs were treated as single sequences for training. And finally, for Gemma, specially formatted prompt templates were applied, including instructions and expected outputs.

**Model Fine-tuning Configurations.** We fine-tuned three different language models with distinct architectures and capabilities. Table 1 summarizes the key configurations used for each model.

Table 1: Fine-tuning configurations across language models

| Configuration | T5 | GPT-2 | Gemma |
|---|---|---|---|
| Parameters | 60M | 117M | 2B |
| Learning rate | $5 \times 10^{-5}$ | $5 \times 10^{-4}$ | $3 \times 10^{-5}$ |
| Batch size | 4 | 4 | 2+4 accum. |
| Training steps | 800 | 600-1000 | 336 |
| LR schedule | Linear | Linear | Cosine |
| Efficient tuning | No | No | LoRA ($r$=16) |

We adapted each model specifically for our morphological assessment task. For T5, we structured inputs as metadata-rich prompts with answer numbers as targets. GPT-2 was trained on complete question-answer sequences using the EOS token for padding. Gemma required more efficient approaches—we applied LoRA to attention modules, reducing trainable parameters to just 0.147% (3.7M) while maintaining quality. This parameter-efficient approach was particularly important given Gemma's size, allowing us to fine-tune on a single NVIDIA A100 GPU. For all models, we employed the AdamW optimizer with model-specific learning rates and schedules as shown in the table.

### 5.1.2 Large Language Model (LLM).

In parallel, we employ GPT-3.5 (approximately 175B parameters[2]); as a larger-scale, pre-trained model. Unlike the SLMs, GPT-3.5 is used in an *off-the-shelf* manner without additional fine-tuning, relying on prompt design alone to steer its generative process.

Through this comprehensive approach, we were able to systematically compare the performance of small and large language models in morphological assessment item generation, while evaluating how different prompting strategies help bridge the performance gap between them.

### 5.2 Fold Two: Prompting Techniques

Prompting guides a language model's text generation through task descriptions or examples. In our setting, prompts help models internalize morphological goals—such as identifying prefixes, suffixes, and root words—while aligning distractors and difficulty with K–12 expectations.

We use several strategies. In **zero-shot prompting**, the model receives only a task instruction (e.g., "Generate a Grade 3 question on the prefix un-") and must infer the format independently. **Few-shot prompting** adds two or three sample Q&A pairs to demonstrate structure and tone. With **chain-of-thought (CoT)** prompting, the model explains its reasoning (e.g., why a prefix applies to a word or how distractors are constructed). **Role-based prompting** extends CoT by simulating perspectives: a teacher checks grade-level suitability, a student assesses clarity and challenge, and a psychometrician validates morphological focus. Lastly, **sequential prompting** breaks generation into stages. In the **one-pass** version, the model performs all steps in a single response. In the **multi-step** variant, it selects a word, drafts a question, and refines distractors across separate calls.

These strategies enable both small and mid-sized models to produce morphologically accurate, grade-appropriate MCQs with strong educational alignment.

## 6 Evaluation

Item quality assessment is carried out through automatic NLP-based and expert-informed evaluations. First, automatic evaluation metrics are reviewed, and then we proceed with the human expert evaluation, followed by the state-of-the-art (SOTA) LLM-based evaluations (GPT 4.1).

### 6.1 Automatic Evaluation Metrics

We employed a suite of automated metrics to evaluate multiple-choice items across four dimensions: grammar, complexity, readability, and fluency. **Grammar** was assessed using `language_tool_python` to check spelling, punctuation, and syntax, yielding a grammar score of $1 - \frac{\text{error\_density}}{10}$, scaled to a 0–100 range. **Complexity** was measured with `spaCy` for lexical and structural sophistication, including average word length, sentence length, vocabulary diversity (unique word ratio), POS distribution, and syntactic tree depth; the complexity score is based on normalized syntactic depth. **Readability** was computed using the Flesch Reading Ease (FRE) formula, which evaluates ease of comprehension, with the final readability score obtained by dividing the FRE score by 100. **Fluency** employed GPT-2 perplexity as a proxy for naturalness and coherence, with a fluency score of $1 - \frac{(\text{perplexity} - 20)}{100}$, clipped between 0 and 1, and then scaled to 0–100.

### 6.2 Human Expert Evaluation followed by GPT-4.1

In addition to automatic metrics, we conducted a human evaluation on a subset of generated items from different question types, focusing on zero-shot output (the weakest prompting method) to allow more comprehensive assessments. Experts scored each item on five binary dimensions, each reflecting a key aspect of morphological assessment validity. **Instruction Clarity** ensures the item's wording is unambiguous, while **Accuracy of the**

---

[2]The 175B parameter figure is often attributed to GPT-3; GPT-3.5 specifics remain approximate due to limited official parameter disclosures.

**Correct Answer** checks that the designated correct choice is unequivocally valid, without alternative interpretations. **Quality of Distractors** examines whether incorrect options create appropriate cognitive challenge by resembling the correct answer yet remain clearly wrong upon reasoning. **Word Difficulty Appropriateness** evaluates if the model-assigned difficulty level aligns with the familiarity and frequency of affixes and roots and the complexity of the morphological structure. **Task Difficulty Alignment** verifies whether the assigned difficulty reflects the cognitive demands of the item, from simple recognition tasks to more complex transformations involving phonological or orthographic manipulation. Expert-labeled items serve as reference points for GPT-4.1, a high-capacity language model used to simulate expert evaluation and ensure consistent large-scale scoring.

## 7 Results

The results are divided into two main folds: (a) automatic evaluations using NLP-based techniques and (b) human expert evaluations, followed by GPT-4.1 evaluations based on expert-labeled examples and criteria. Before reviewing the results, database challenges should be noted.

### 7.1 Database Constraints and Model Selection

Throughout our experiments, two major challenges emerged. First, our dataset comprised only 268 items spanning 13 question types. Fine-tuning large-scale models such as GPT-2 and T5 typically requires substantially larger and more diverse data; in our setting, these models struggled to adapt beyond the limited training examples and failed to generate sufficiently varied or context-sensitive questions. Second, attempts to fine-tune these models yielded underwhelming performance, as they tended to overfit and produce repetitive or semantically bland items. Consequently, the final comparison focuses on two models: **(a) Gemma (2B parameters)**, fine-tuned on the 268-item dataset, and **(b) GPT-3.5 (175B parameters)**, used off-the-shelf without fine-tuning. This scenario highlights the need for larger and more representative datasets.

### 7.2 Automatic Evaluations

We evaluate the generated items using four automated metrics: Grammar, Complexity, Readability, and Fluency—and test both Gemma (fine-tuned) and GPT-3.5 under six prompting strategies (see Table 2).

Overall, **GPT-3.5** outperforms **Gemma** in zero-shot prompting, especially on **Fluency** (33.93 vs. 31.64) and **Grammar** (97.62 vs. 90.19). However, Gemma's performance improves considerably with more structured prompts. **Chain-of-Thought (CoT)** boosts Gemma's Fluency from 31.64 to 36.28 and Readability from 74.68 to 77.65, indicating that step-by-step reasoning enhances coherence and clarity. **Role Conditioning** adds further gains; under CoT + Role, Gemma's Grammar jumps from 90.19 to 94.86, suggesting that domain-specific framing (e.g., "Teacher" or "Psychometrician") refines syntactic correctness. **Sequential (Multi-Step) Prompting** yields Gemma's highest Grammar score (95.64) and second-highest Complexity (91.01), though increases in Fluency (29.83) and Readability (75.45) are more modest, indicating that breaking the task into finer steps contributes to structurally sound questions.

In summary, the table shows that while GPT-3.5 excels out-of-the-box, smaller models like Gemma can approach comparable quality when guided by carefully designed, multi-step prompts. Since some automatic metrics may not fully capture quality variation—particularly for certain question types—detailed per-category results are provided in the supplementary material.

### 7.3 Human Expert Evaluations, followed by GPT-4.1

In the second fold, we conducted human assessments on a subset of zero-shot items. Each item was rated along five binary dimensions—Instruction Clarity, Accuracy of the Correct Answer, Quality of Distractors, Word Difficulty Appropriateness, and Task Difficulty Alignment.

Table 18 (provided in Appendix D) summarizes the average human evaluation scores across five dimensions for a subset of 13 question types. Overall, items assessing affix identification and basic derivational knowledge (QT1–QT5, QT7, QT10) received relatively high scores (≥3.2 out of 5), particularly in terms of instruction clarity and answer accuracy. In contrast, items requiring semantic reasoning (QT6, QT11–QT13) performed poorly, often due to vague task definitions, misleading distractors, or unrealistic task framing. Common weaknesses included overestimated difficulty ratings, limited affix variety (e.g., overuse of "un-"), and insufficient morphemic structure in target words. These findings highlight specific areas where zero-shot item generation could be improved.

Table 2: Automatic Evaluation of Prompting Strategies for Gemma (2B) and GPT-3.5 (175B). Scores are scaled to 0–100.

| Prompting Strategy | Complexity | | Fluency | | Grammar | | Readability | |
| --- | --- | --- | --- | --- | --- | --- | --- | --- |
| | Gemma | GPT-3.5 | Gemma | GPT-3.5 | Gemma | GPT-3.5 | Gemma | GPT-3.5 |
| Zero-Shot | 90.67 | 93.23 | 31.64 | 33.93 | 90.19 | 97.62 | 74.68 | 76.52 |
| Few-Shot | 91.10 | 93.40 | 32.10 | 41.46 | 91.76 | 97.76 | 76.01 | 79.68 |
| CoT | 91.54 | 92.46 | 36.28 | 36.97 | 91.13 | 94.52 | 77.65 | 78.44 |
| CoT + Role | 90.03 | 91.03 | 29.58 | 34.43 | 94.86 | 91.93 | 76.24 | 82.23 |
| CoT + Sequential (One-Go) | 90.71 | 91.78 | 30.73 | 35.94 | 94.62 | 96.29 | 76.24 | 79.58 |
| CoT + Sequential (Multi-Step) | 91.01 | 93.15 | 29.83 | 35.43 | 95.64 | 93.12 | 75.45 | 75.83 |
| Average | 90.87 | 92.51 | 31.75 | 35.70 | 93.13 | 94.58 | 76.25 | 78.41 |

These expert-annotated items were then used to *condition GPT-4.1* to emulate expert judgment. Specifically, GPT-4.1 was prompted with examples of human-labeled outputs and instructed to assign scores in the same five dimensions for all generated items from both Gemma and GPT-3.5. This approach provides a more pedagogically informed view of item quality than raw automated metrics, particularly for morphological correctness and nuanced difficulty alignment.

Table 3 reports the consolidated scores for both Gemma (fine-tuned) and GPT-3.5 (off-the-shelf) under six prompting strategies. Unlike the automated metrics, these dimension-level scores directly capture pedagogical appropriateness (e.g., whether the question accurately reflects morphological complexity and grade-level alignment).

**Analysis of Human+GPT-4.1 Results.** **Zero-Shot** serves as a baseline for both models, where Gemma achieves a total score of 3.94 vs. GPT-3.5's 3.34. Although GPT-3.5 excelled in the automated Grammar and Fluency metrics from Section 7.2, human-based scoring reveals that Gemma's fine-tuning yields stronger alignment with morphological goals out of the gate. **Few-Shot** prompts boost both models by about 0.15–0.30 in total score.

Moving to advanced prompting, **CoT + Seq** emerges as the top performer for both models, with Gemma reaching 4.08 and GPT-3.5 reaching 3.55 in total score. This reaffirms the importance of step-by-step reasoning combined with a structured prompt flow. Interestingly, **CoT + Seq_Rl** (role-based, multi-step) underperforms for GPT-3.5 (2.87 total score) but remains high for Gemma (4.01). This discrepancy suggests that certain multi-role instructions may require more careful tuning for large models, whereas Gemma—in part due to extensive fine-tuning—interprets and integrates these role prompts effectively.

**Comparison with Automated Metrics.** Cross-referencing Table 3 against the automated results in Table 2 yields two salient observations:

- **Differences in Model Ranking:** Although GPT-3.5 led most categories under automated evaluations, Gemma obtains higher holistic scores (3.90+ vs. 3.34 for GPT-3.5 on average) from human+GPT-4.1. This indicates that while GPT-3.5 excels at producing grammatically refined text, Gemma's fine-tuned approach better aligns with morphological correctness and K–12 item design requirements.

- **Value of Step-by-Step Prompts:** Both sets of results converge on the efficacy of multi-step chain-of-thought prompting. Automated scores highlighted improvements in Fluency and Grammar, whereas human-based scores confirm more appropriate distractors, word choices, and difficulty alignments.

Overall, these expert-informed findings underscore the importance of not solely relying on surface-level text metrics. Morphological accuracy, instructional clarity, and grade-level alignment often diverge from purely linguistic measures of quality, reinforcing the need for combined automated and human+LLM evaluation pipelines in developing robust educational content.

# 8 Discussion

Our results reveal how different scales of language models and prompting strategies impact the quality of automatically generated morphological items for K–12 education. Below, we synthesize key insights and discuss implications for both research and practice.

Table 3: Human + GPT-4.1 evaluation of **Gemma** versus **GPT-3.5**. Each sub-score ranges from 0–1; the *Total* ranges from 0–5.

| Prompting Strategy | Clarity of Instruction | | Accuracy of Correct Answer | | Quality of Distractors | | Word Difficulty | | Task Difficulty | | Total (0–5) | |
|---|---|---|---|---|---|---|---|---|---|---|---|---|
| | **Gem.** | **GPT** | **Gem.** | **GPT** | **Gem.** | **GPT** | **Gem.** | **GPT** | **Gem.** | **GPT** | **Gem.** | **GPT** |
| Zero-Shot | 0.9417 | 0.8356 | 0.9061 | 0.7003 | 0.7961 | 0.7243 | 0.8026 | 0.6455 | 0.4854 | 0.4247 | 3.9353 | 3.3356 |
| Few-Shot | 0.9421 | 0.8530 | 0.8994 | 0.6991 | 0.7927 | 0.7333 | 0.8049 | 0.7043 | 0.4451 | 0.4821 | 3.8933 | 3.4821 |
| CoT | 0.8747 | 0.8393 | 0.8031 | 0.7128 | 0.6854 | 0.7265 | 0.8056 | 0.6496 | 0.4041 | 0.4137 | 3.5831 | 3.3470 |
| CoT + Role | 0.9544 | 0.8731 | 0.9088 | 0.7702 | 0.7872 | 0.7307 | 0.8237 | 0.6432 | 0.4407 | 0.4443 | 3.9179 | 3.4717 |
| CoT + Seq | 0.9745 | 0.8638 | 0.9209 | 0.7500 | 0.8163 | 0.6845 | 0.8444 | 0.7276 | 0.5204 | 0.5103 | 4.0816 | 3.5483 |
| CoT + Seq_Rl | 0.9454 | 0.6781 | 0.9156 | 0.5856 | 0.8015 | 0.6164 | 0.8263 | 0.6096 | 0.5037 | 0.3716 | 4.0074 | 2.8733 |
| **Grand Avg.** | 0.9382 | 0.8238 | 0.8913 | 0.7029 | 0.7788 | 0.7027 | 0.8188 | 0.6632 | 0.4675 | 0.4410 | 3.9024 | 3.3428 |

**Model Scale, Data Constraints, and Baseline Performance.** From the outset, our inability to effectively fine-tune T5 and GPT-2 underscores the limitations of a small (268-item) dataset. Large-scale models generally demand extensive data to avoid underfitting or repetitive outputs. By contrast, Gemma—although still a *small* model relative to 175B-parameter LLMs—was sufficiently adaptable via parameter-efficient fine-tuning. GPT-3.5, with a much larger pretraining corpus, excelled off-the-shelf, reflecting its broad knowledge base. These findings highlight an important trade-off: *without a robust dataset, smaller models struggle, whereas off-the-shelf large models can maintain solid performance even in data-constrained contexts.*

**Prompting Strategies Can Bridge Performance Gaps.** Despite GPT-3.5's initially stronger showing in automated metrics (Grammar, Fluency), advanced prompting significantly boosted Gemma's quality scores, both automatically and under human+GPT-4.1 evaluations. In particular, *Chain-of-Thought (CoT) plus Sequential* approaches yielded Gemma's highest total scores, validating a stepwise reasoning structure that aligns with morphological item creation (e.g., breaking a word into affixes, specifying reading level, then generating distractors). Although GPT-3.5 remained competitive across most metrics, the role-based sequential prompts (*CoT + Seq_Rl*) performed inconsistently, highlighting that complex role instructions may require careful calibration for large-scale LLMs.

**Discrepancies Between Automated and Human+LLM Evaluations.** A salient outcome is that automated evaluations favored GPT-3.5, yet human+GPT-4.1 scoring often favored Gemma, especially in zero-shot or few-shot cases. This discrepancy points to fundamental differences in what each metric values: automated methods prioritize linguistic coherence and syntactic correctness, whereas expert-informed scores emphasize morphological alignment, question clarity, and appropriate difficulty levels. For educational item generation, the latter factors are crucial, suggesting that purely linguistic metrics may be insufficient for evaluating domain-specific constructs like morphology.

**Morphological Complexity and Construct Validity.** Morphologically focused MCQs require precise handling of prefixes, suffixes, and roots. Our experiments confirm that misaligned distractors or incorrect word breakdowns can severely degrade the educational value of an item. Prompting strategies that explicitly guide the model to consider morphological details—whether via role-based instructions (*Teacher*, *Psychometrician*) or a structured chain-of-thought—consistently mitigated such errors. These findings support the growing consensus that prompt design should incorporate domain knowledge to maintain construct validity (Wei et al., 2022; Yao et al., 2023).

**Implications for Real-World Deployment.** In practice, deploying GPT-3.5 (or similar LLMs) can be both compute- and cost-intensive, while smaller models require more elaborate prompts or data augmentation to produce robust educational content. Our observations suggest two practical workflows:

1. **Large Model, Minimal Tuning:** For immediate, relatively high-quality item generation, an off-the-shelf LLM can yield decent outcomes but may lack domain-specific nuance without advanced prompting techniques.

2. **Smaller Model, Deep Prompting/Fine-Tuning:** A mid-scale model like Gemma can outperform the large model on domain-oriented scores if fine-tuned and guided by carefully designed multi-step prompts, requiring lower long-term operational costs.

Both routes benefit from incorporating a *human-in-the-loop* or LLM-assisted review phase, which remains vital for ensuring appropriate difficulty alignment, morphological accuracy, and minimal distractor confusion.

**Toward More Comprehensive Evaluations.** Our dual approach (automatic metrics plus expert/GPT-4.1 evaluations) highlights the importance of going beyond surface-level text quality. Future research could refine automated metrics to capture morphological correctness—e.g., using morphological parsers or specialized psychometric indicators—while also exploring advanced prompt-chaining methods (e.g., iterative self-refinement (Madaan et al., 2023)) to further enhance item generation. Given the domain-specific nuances in K–12 contexts, such multi-pronged evaluations are likely to become the norm for robust AIG pipelines.

**Concluding Remarks.** Overall, this study underscores the interplay between model scale, data constraints, and prompt engineering in generating high-quality morphological items. While large-scale LLMs can deliver strong baseline performance even with minimal data, fine-tuned mid-scale models—guided by chain-of-thought prompts—can surpass them in domain alignment. Bridging automated and expert-based evaluation methodologies thus emerges as a key priority, ensuring that future AIG systems meet both linguistic and pedagogical standards in K–12 educational settings.

## Future Directions

While the present study demonstrates the efficacy of prompt-driven, mid-scale fine-tuning alongside large, off-the-shelf models, several promising extensions remain.

First, integrating *prompt optimization and search* techniques (e.g., Tree-of-Thoughts (Yao et al., 2023) or iterative self-refinement (Madaan et al., 2023)) could further enhance morphological accuracy and item difficulty calibration by systematically exploring variations in chain-of-thought or role assignments.

Second, the value of *larger or synthetic datasets* becomes clear, as data augmentation through paraphrasing or minor morphological manipulations could expand coverage of rarely used affixes and improve generalization for mid-scale models beyond limited in-domain samples.

Third, a *real-time evaluation* incorporating partial feedback from educators or automated morphological parsers may make generation more adaptive, enabling immediate checks of affix consistency or plausible distractors during item drafting.

Lastly, *diverse assessment types* (e.g., reading comprehension, domain-specific vocabulary) may also benefit from these techniques, although each domain's unique constraints—such as scientific terminology or complex grammatical rules—require further investigation.

## Limitations

Although our results demonstrate the feasibility of generating morphological multiple-choice items with both small and large language models, several limitations must be acknowledged. Our dataset contains only 268 items across 13 morphological question types, which may not capture all K–12 morphological nuances. This small size hinders effective fine-tuning for mid-to-large models, limiting generalization beyond the specific domain or linguistic constructs.

Our approach focuses primarily on morphological knowledge and does not fully encompass other K–12 literacy skills. Subtle aspects—such as age-appropriate vocabulary or question-stem complexity—remain incompletely assessed by automated metrics, so human review is essential to ensure alignment with grade-level objectives and learning standards.

Performance improvements largely depend on careful prompt engineering, which demands expertise in both linguistics and model behavior. Inconsistent or poorly crafted prompts can lead to erratic generations that undermine the reliability of automated item creation.

The chosen metrics (Grammar, Complexity, Readability, Fluency) reflect broad text quality but do not guarantee accurate morphological breakdowns or educational soundness. A model might score high linguistically yet fail to produce valid morphological contrasts or plausible distractors. Although LLM-based "expert simulation" (e.g., GPT-4.1) can approximate human judgment at scale, it may overlook nuanced educational standards or edge cases in specialized morphological content.

Although large models (e.g., GPT-3.5) provide strong out-of-the-box performance, they can be expensive and subject to API constraints, making them less accessible in resource-limited contexts. Smaller models require extensive tuning and advanced prompting to match this performance, thus increasing computational and time costs.

Finally, our study targets elementary-level morphological MCQs, and the methods may not directly translate to domains like reading comprehension or mathematics, which have different structures and cognitive demands. Despite these constraints, our work offers a starting point for using language models in automated assessment generation, emphasizing the importance of prompt design and model selection. Future research should expand datasets, refine evaluation protocols, and incorporate richer feedback loops to build a more comprehensive, scalable pipeline.

# Appendix

## A  Detailed Question Types

Table 4 below outlines the 13 distinct question types referenced in Section 4:

Table 4: Summary of Question Types (QT1–QT13).

| Question Type | Description |
|---|---|
| **QT1** | Identify the *prefix* in a word from three choices. |
| **QT2** | Identify the *suffix* in a word from three choices. |
| **QT3** | Identify the *root word* in a word from three choices. |
| **QT4** | Choose the word that does *not share the same prefix* as the others. |
| **QT5** | Choose the word that does *not share the same suffix* as the others. |
| **QT6** | Choose the correct *transformed word* based on a meaning shift. |
| **QT7** | Select the correct *meaning of an affixed word* from three choices. |
| **QT8** | Choose the correct *spelling of a word with a suffix*, avoiding two misspellings. |
| **QT9** | Break a word into its *prefix, root, and suffix* from three given segmentations. |
| **QT10** | Select the correct *definition of the prefix* in a word. |
| **QT11** | Select the correct *definition of the root word* in an affixed word. |
| **QT12** | Select the correct *definition or function of the suffix* in a word. |
| **QT13** | Determine the *meaning of a complex word* based on its morphemes. |

## B  Example Items for Each Question Type

**QT1.** What is the prefix in the word *miswrote*?
A. mis
B. misw
C. ote
**Answer:** A

**QT2.** What is the suffix in the word *governmental*?
A. al
B. mental
C. government
**Answer:** A

**QT3.** What is the root word in *subspecialty*?
A. special
B. specialty
C. species
**Answer:** A

**QT4.** Find the word that does **NOT** have the same prefix as the other two words.
A. unique
B. unexcitable
C. unkindness
**Answer:** A

**QT5.** Find the word that does **NOT** have the same suffix as the other two words.
A. uniquely
B. ugly
C. usefully
**Answer:** B

**QT6.** Change the word *undivided* to mean "smaller parts of a whole."
A. disdivide
B. antidivision
C. subdivisions
**Answer:** C

**QT7.** What is the meaning of the word *nonperishable*?
A. fresh; cannot be stored
B. made to stay good while being stored
C. afraid of going bad
**Answer:** B

**QT8.** Select the correctly spelled word.
A. prepublicashun
B. prepublishation
C. prepublication
**Answer:** C

**QT9.** Break the word *rehammering* into parts based on prefixes, roots, and suffixes.
A. re/hammering
B. re/hammer/ing
C. re/ham/mering
**Answer:** B

**QT10.** What is the meaning of the prefix in *transplant*?
A. within
B. movement between
C. different from
**Answer:** B

**QT11.** What is the meaning of the root word in *overtraining*?
A. to learn through practice
B. a vehicle that uses railroads
C. to injure by too much work
**Answer:** A

**QT12.** What is the meaning of the suffix in *expressive*?
A. person who
B. relating to
C. the action of
**Answer:** B

**QT13.** If *district* refers to an area, what does *subdistricts* mean?
A. larger parts of an area
B. smaller parts of an area
C. a few different areas
**Answer:** B

## C  Detailed Evaluation Tables

This supplementary material presents detailed evaluation results for each question type assessed in the study. For every question, the performance of two language models—Gemma and GPT—was evaluated under various prompting strategies. The evaluation metrics include Grammar, Complexity, Readability, and Fluency scores. Tables 5 through 17 summarize these results.

While the evaluation metrics provide a structured comparison across question types, some metrics vary in their informativeness depending on the nature of the task. Below we provide intuitive interpretations and caveats for the main metrics used.

- **Fluency.** The fluency score generally scales with response length, making it less meaningful for short, discrete-response question types like QT1–QT3, QT6, QT8, and QT10–QT12. In contrast, for more expressive and context-rich tasks (QT4, QT5, QT7, QT13), where responses require explanation or contrastive reasoning, fluency becomes a more relevant indicator of natural language flow.

- **Grammar.** Most question types involve grammatically simple structures. However, QT12 presents a challenge: it contains responses that include partial or bound morphemes (e.g., suffixes in isolation), which automated grammar scoring tools may erroneously flag as ungrammatical. As such, grammar scores in QT12 are not always reliable indicators of model quality.

- **Readability.** This metric captures how easily a human reader can process a generated response. Higher scores often reflect simpler vocabulary and sentence structure. For question types involving affix meanings or morphological decompositions (e.g., QT9–QT13), readability scores may naturally be lower due to the technical nature of the vocabulary and the need for precise linguistic explanation. Conversely, QT1–QT3 and QT6–QT8, which involve direct selection or spelling tasks, usually yield clearer and more readable outputs.

- **Complexity.** This metric reflects the syntactic and lexical sophistication of a model's output. It becomes most informative in questions that allow for elaboration or require reasoning—particularly QT4, QT5, QT7, and QT13—where a more nuanced answer may indicate deeper understanding. For direct identification or spelling tasks (QT1–QT3, QT6, QT8), high complexity may actually suggest unnecessary verbosity or over-generation, and thus a lower complexity score may indicate more appropriate performance.

These observations highlight the importance of interpreting automated metrics in the context of question type. Direct comparisons across question types should be made with caution, and metric-specific limitations should be considered when drawing conclusions about model performance.

Table 5: Evaluation Results for question type 1

| Prompting Strategy | Complexity | | Fluency | | Grammar | | Readability | |
|---|---|---|---|---|---|---|---|---|
| | Gemma | GPT-3.5 | Gemma | GPT-3.5 | Gemma | GPT-3.5 | Gemma | GPT-3.5 |
| Zero-Shot | 80 | 80 | 0 | 1.10 | 75 | 100 | 70.32 | 74.79 |
| Few-Shot | 80 | 80 | 0.53 | 0.07 | 84.37 | 100 | 73.45 | 77.92 |
| CoT | 80 | 80 | 0 | 0 | 90.62 | 100 | 72.08 | 78.55 |
| CoT + RC | 80 | 80.44 | 0 | 0 | 93.94 | 100 | 74.70 | 79.66 |
| CoT + Sequential | 80 | 79.56 | 0 | 0.12 | 84.85 | 100 | 75.55 | 79.80 |
| CoT + Sequential_Rl | 80.59 | 80 | 0.63 | 0.59 | 91.18 | 100 | 69.52 | 77.25 |
| Average | 80.10 | 80 | 0.20 | 0.31 | 86.66 | 100 | 72.60 | 77.99 |

Table 6: Evaluation Results for question type 2

| Prompting Strategy | Complexity | | Fluency | | Grammar | | Readability | |
|---|---|---|---|---|---|---|---|---|
| | Gemma | GPT-3.5 | Gemma | GPT-3.5 | Gemma | GPT-3.5 | Gemma | GPT-3.5 |
| Zero-Shot | 80 | 80 | 2.01 | 0 | 90.48 | 100 | 75.19 | 74.16 |
| Few-Shot | 80 | 80 | 0 | 0 | 91.67 | 97.78 | 75.61 | 73.85 |
| CoT | 82.53 | 80 | 0 | 0 | 93.42 | 100 | 75.76 | 76.31 |
| CoT + RC | 80 | 80 | 1.69 | 0 | 92 | 100 | 77.43 | 76.04 |
| CoT + Sequential | 80 | 80 | 4.23 | 0 | 96.67 | 97.78 | 74.32 | 75.25 |
| CoT + Sequential_Rl | 80 | 80 | 2.82 | 0 | 96.67 | 100 | 74.79 | 78.99 |
| Average | 80.42 | 80 | 1.80 | 0 | 93.49 | 99.26 | 75.51 | 75.77 |

Table 7: Evaluation Results for question type 3

| Prompting Strategy | Complexity | | Fluency | | Grammar | | Readability | |
|---|---|---|---|---|---|---|---|---|
| | Gemma | GPT-3.5 | Gemma | GPT-3.5 | Gemma | GPT-3.5 | Gemma | GPT-3.5 |
| Zero-Shot | 81.6 | 79.56 | 22.06 | 1.20 | 76 | 100 | 80.70 | 88.96 |
| Few-Shot | 80 | 79.56 | 30.96 | 4.61 | 88 | 100 | 76.50 | 85.02 |
| CoT | 80.74 | 80 | 22.48 | 2.20 | 74.07 | 100 | 80.85 | 87.31 |
| CoT + RC | 80 | 79.56 | 27.16 | 3.22 | 92.31 | 100 | 79.65 | 89.63 |
| CoT + Sequential | 80 | 80 | 19.95 | 2.66 | 93.33 | 100 | 79.35 | 87.10 |
| CoT + Sequential_Rl | 80 | 78.67 | 25.30 | 4.40 | 80 | 100 | 80.90 | 87.89 |
| Average | 80.39 | 79.56 | 24.65 | 3.05 | 83.95 | 100 | 79.66 | 87.65 |

Table 8: Evaluation Results for question type 4

| Prompting Strategy | Complexity | | Fluency | | Grammar | | Readability | |
|---|---|---|---|---|---|---|---|---|
| | Gemma | GPT-3.5 | Gemma | GPT-3.5 | Gemma | GPT-3.5 | Gemma | GPT-3.5 |
| Zero-Shot | 98.62 | 100 | 90.80 | 82.74 | 100 | 100 | 92.82 | 93.85 |
| Few-Shot | 100 | 100 | 94.17 | 98.60 | 88.89 | 100 | 95.14 | 95.73 |
| CoT | 99.55 | 100 | 94.41 | 79.78 | 97.79 | 100 | 93.81 | 95.69 |
| CoT + RC | 100 | 99.11 | 98.60 | 80.15 | 100 | 100 | 95.73 | 95.26 |
| CoT + Sequential | 100 | 100 | 98.60 | 86.89 | 100 | 100 | 95.73 | 91.31 |
| CoT + Sequential_Rl | 100 | 100 | 98.60 | 88.70 | 100 | 100 | 95.73 | 87.84 |
| Average | 99.70 | 99.85 | 95.86 | 86.14 | 97.78 | 100 | 94.83 | 93.28 |

Table 9: Evaluation Results for question type 5

| Prompting Strategy | Complexity | | Fluency | | Grammar | | Readability | |
|---|---|---|---|---|---|---|---|---|
| | Gemma | GPT-3.5 | Gemma | GPT-3.5 | Gemma | GPT-3.5 | Gemma | GPT-3.5 |
| Zero-Shot | 99.29 | 100 | 94.58 | 76.37 | 100 | 100 | 94.44 | 95.43 |
| Few-Shot | 100 | 100 | 97.21 | 98.09 | 97.54 | 100 | 95.11 | 95.73 |
| CoT | 100 | 99.56 | 95.44 | 75.75 | 98.72 | 100 | 93.93 | 95.69 |
| CoT + RC | 100 | 100 | 96.72 | 76.37 | 96.15 | 100 | 94.86 | 95.43 |
| CoT + Sequential | 100 | 100 | 98.09 | 80.94 | 100 | 100 | 95.73 | 92.80 |
| CoT + Sequential_Rl | 100 | 100 | 98.09 | 80.80 | 100 | 100 | 95.73 | 92.80 |
| Average | 99.88 | 99.93 | 96.69 | 81.39 | 98.74 | 100 | 94.97 | 94.65 |

Table 10: Evaluation Results for question type 6

| Prompting Strategy | Complexity | | Fluency | | Grammar | | Readability | |
|---|---|---|---|---|---|---|---|---|
| | Gemma | GPT-3.5 | Gemma | GPT-3.5 | Gemma | GPT-3.5 | Gemma | GPT-3.5 |
| Zero-Shot | 100 | 92 | 23.78 | 4.03 | 96.75 | 97.78 | 77.21 | 73.45 |
| Few-Shot | 97.50 | 98.22 | 23.55 | 24.44 | 100 | 100 | 84.28 | 87.78 |
| CoT | 99.26 | 84.89 | 20.49 | 10.16 | 100 | 100 | 74.78 | 86.30 |
| CoT + RC | 97.50 | 99.11 | 19.42 | 28.77 | 100 | 90.04 | 86.02 | 89.34 |
| CoT + Sequential | 98 | 97.78 | 24.56 | 26.78 | 100 | 96.90 | 84.40 | 79.76 |
| CoT + Sequential_Rl | 99.33 | 96.89 | 11.08 | 20.31 | 100 | 99.18 | 77.52 | 69.26 |
| Average | 98.60 | 94.82 | 20.48 | 19.08 | 99.46 | 97.31 | 80.70 | 80.98 |

Table 11: Evaluation Results for question type 7

| Prompting Strategy | Complexity | | Fluency | | Grammar | | Readability | |
|---|---|---|---|---|---|---|---|---|
| | Gemma | GPT-3.5 | Gemma | GPT-3.5 | Gemma | GPT-3.5 | Gemma | GPT-3.5 |
| Zero-Shot | 80 | 92.44 | 57.70 | 73.96 | 100 | 98.15 | 67.10 | 83.39 |
| Few-Shot | 80 | 99.11 | 52.85 | 82.91 | 100 | 95.56 | 68.94 | 77.65 |
| CoT | 80 | 92.44 | 55.56 | 73.81 | 100 | 100 | 74.33 | 82.74 |
| CoT + RC | 80 | 76.44 | 61.07 | 66.29 | 100 | 97.78 | 65.03 | 89.62 |
| CoT + Sequential | 80 | 88 | 59.28 | 64.45 | 100 | 94.25 | 65.39 | 84.60 |
| CoT + Sequential_Rl | 80 | 99.11 | 63.90 | 74.13 | 100 | 96.98 | 67.27 | 72.83 |
| Average | 80 | 91.26 | 58.39 | 72.59 | 100 | 97.12 | 68.01 | 81.80 |

Table 12: Evaluation Results for question type 8

| Prompting Strategy | Complexity | | Fluency | | Grammar | | Readability | |
|---|---|---|---|---|---|---|---|---|
| | Gemma | GPT-3.5 | Gemma | GPT-3.5 | Gemma | GPT-3.5 | Gemma | GPT-3.5 |
| Zero-Shot | 80 | 98.22 | 0 | 43.80 | 100 | 98.41 | 49.48 | 65.33 |
| Few-Shot | 80 | 80 | 0 | 0 | 100 | 100 | 49.48 | 49.48 |
| CoT | 81.60 | 100 | 5.94 | 58.38 | 100 | 97.11 | 52.03 | 71.01 |
| CoT + RC | 80 | 95.56 | 0 | 42.77 | 100 | 76.21 | 49.48 | 70.75 |
| CoT + Sequential | 80 | 98.22 | 0 | 62.43 | 100 | 93.60 | 49.48 | 67.68 |
| CoT + Sequential_Rl | 80 | 100 | 0 | 61.17 | 100 | 88.06 | 49.48 | 68.21 |
| Average | 80.27 | 95.33 | 0.99 | 44.76 | 100 | 92.23 | 49.90 | 65.41 |

Table 13: Evaluation Results for question type 9

| Prompting Strategy | Complexity | | Fluency | | Grammar | | Readability | |
|---|---|---|---|---|---|---|---|---|
| | Gemma | GPT-3.5 | Gemma | GPT-3.5 | Gemma | GPT-3.5 | Gemma | GPT-3.5 |
| Zero-Shot | 99.13 | 98.67 | 35.83 | 11.04 | 92.31 | 100 | 57.71 | 54.27 |
| Few-Shot | 100 | 100 | 44.69 | 74.96 | 86.62 | 100 | 53.30 | 67.76 |
| CoT | 99.29 | 93.33 | 51.09 | 5.95 | 91.80 | 96.27 | 57.02 | 63.56 |
| CoT + RC | 100 | 92.89 | 46.87 | 12.35 | 100 | 97.78 | 54.12 | 59.65 |
| CoT + Sequential | 100 | 91.11 | 48.90 | 4.67 | 94.69 | 100 | 54.51 | 68.33 |
| CoT + Sequential_Rl | 100 | 91.56 | 43.85 | 5.14 | 97.44 | 95.76 | 56.54 | 69.07 |
| Average | 99.74 | 94.59 | 45.21 | 19.02 | 93.81 | 98.30 | 55.53 | 63.77 |

Table 14: Evaluation Results for question type 10

| Prompting Strategy | Complexity | | Fluency | | Grammar | | Readability | |
|---|---|---|---|---|---|---|---|---|
| | Gemma | GPT-3.5 | Gemma | GPT-3.5 | Gemma | GPT-3.5 | Gemma | GPT-3.5 |
| Zero-Shot | 99.05 | 92.44 | 1.24 | 16.87 | 100 | 84.62 | 82.11 | 73.50 |
| Few-Shot | 100 | 97.33 | 2.93 | 2.92 | 100 | 88.89 | 78.87 | 80.22 |
| CoT | 97.60 | 92.67 | 2.50 | 7.72 | 100 | 72.79 | 81.73 | 74.84 |
| CoT + RC | 96.67 | 89.56 | 0 | 15.63 | 100 | 76.18 | 82.46 | 82.55 |
| CoT + Sequential | 99.33 | 86.67 | 1.73 | 6.88 | 100 | 63.63 | 81.20 | 82.69 |
| CoT + Sequential_Rl | 97.33 | 87.33 | 0.87 | 8.86 | 100 | 83.48 | 82.36 | 79.88 |
| Average | 98.33 | 91.00 | 1.54 | 9.81 | 100 | 78.27 | 81.45 | 78.95 |

Table 15: Evaluation Results for question type 11

| Prompting Strategy | Complexity | | Fluency | | Grammar | | Readability | |
|---|---|---|---|---|---|---|---|---|
| | Gemma | GPT-3.5 | Gemma | GPT-3.5 | Gemma | GPT-3.5 | Gemma | GPT-3.5 |
| Zero-Shot | 99.20 | 99.11 | 17.30 | 37.57 | 80.00 | 97.98 | 85.44 | 71.03 |
| Few-Shot | 100 | 100 | 18.03 | 76.99 | 82.61 | 100 | 80.40 | 87.83 |
| CoT | 100 | 99.56 | 21.46 | 33.58 | 92.59 | 98.29 | 79.05 | 69.69 |
| CoT + RC | 100 | 96.44 | 6.72 | 38.50 | 91.67 | 100 | 81.77 | 85.88 |
| CoT + Sequential | 100 | 97.27 | 11.92 | 52.74 | 83.33 | 97.93 | 81.77 | 87.87 |
| CoT + Sequential_Rl | 100 | 99.11 | 18.51 | 40.55 | 93.33 | 96.39 | 79.57 | 72.95 |
| Average | 99.87 | 98.58 | 15.65 | 46.66 | 87.26 | 98.43 | 81.33 | 79.21 |

Table 16: Evaluation Results for question type 12

| Prompting Strategy | Complexity | | Fluency | | Grammar | | Readability | |
|---|---|---|---|---|---|---|---|---|
| | Gemma | GPT-3.5 | Gemma | GPT-3.5 | Gemma | GPT-3.5 | Gemma | GPT-3.5 |
| Zero-Shot | 95.45 | 99.56 | 36.15 | 38.85 | 95.45 | 92.22 | 82.10 | 76.19 |
| Few-Shot | 96.52 | 100 | 26.68 | 19.35 | 91.30 | 100 | 82.36 | 79.12 |
| CoT | 96.15 | 99.56 | 28.06 | 20.66 | 76.92 | 64.28 | 81.22 | 82.15 |
| CoT + RC | 92.80 | 95.56 | 33.58 | 20.74 | 92.00 | 57.07 | 82.45 | 81.06 |
| CoT + Sequential | 95.33 | 94.67 | 35.86 | 24.23 | 96.67 | 62.53 | 84.14 | 78.82 |
| CoT + Sequential_Rl | 98.00 | 98.22 | 28.10 | 13.38 | 100.00 | 52.61 | 81.72 | 81.71 |
| Average | 95.71 | 97.93 | 31.40 | 22.87 | 92.06 | 71.45 | 82.33 | 79.84 |

Table 17: Evaluation Results for question type 13

| Prompting Strategy | Complexity | | Fluency | | Grammar | | Readability | |
|---|---|---|---|---|---|---|---|---|
| | Gemma | GPT-3.5 | Gemma | GPT-3.5 | Gemma | GPT-3.5 | Gemma | GPT-3.5 |
| Zero-Shot | 83.81 | 100 | 0 | 53.56 | 67.53 | 100 | 69.96 | 46.52 |
| Few-Shot | 90.40 | 100 | 6.96 | 56.05 | 84.36 | 86.65 | 65.32 | 77.73 |
| CoT | 87.74 | 100 | 12.80 | 60.68 | 64.61 | 100 | 67.92 | 55.94 |
| CoT + RC | 87.20 | 98.67 | 0 | 62.76 | 76.36 | 100 | 67.23 | 74.17 |
| CoT + Sequential | 88.00 | 100 | 0 | 54.78 | 81.41 | 98.52 | 68.93 | 58.76 |
| CoT + Sequential_Rl | 89.33 | 100 | 0 | 62.51 | 85.35 | 98.15 | 70.47 | 47.19 |
| Average | 87.75 | 99.78 | 3.29 | 58.39 | 76.61 | 97.22 | 68.31 | 60.05 |

## D  Human Evaluation Summary

The table below presents average human evaluation scores across five dimensions for a selected subset of all 13 question types. Overall, question types focused on affix recognition and basic derivational skills (QT1–QT5, QT7, QT10) received comparatively high ratings (≥3.2 out of 5), particularly in terms of instruction clarity and correct answer identification. In contrast, question types involving semantic reasoning (QT6, QT11–QT13) showed weaker performance, often due to ambiguous task descriptions, confusing distractors, or implausible task setups. Recurring issues included inflated difficulty scores, repetitive affix use (e.g., frequent use of "un-"), and inadequate morphemic complexity in the target words. These results point to specific areas where zero-shot item generation methods could be improved.

Table 18: Average Human Evaluation Scores for Each Question Type Across Five Dimensions

|      | Clarity of Instruction (0–1) | Quality of Distractors (0–1) | Accuracy of Correct Answer (0–1) | Word Difficulty (0–1) | Task Difficulty (0–1) | Total (out of 5) |
|------|------|------|------|------|------|------|
| QT1  | 1.0 | 0.8 | 0.8 | 0.4 | 0.4 | 3.4 |
| QT2  | 1.0 | 0.6 | 1.0 | 0.8 | 0.6 | 4.0 |
| QT3  | 1.0 | 1.0 | 0.8 | 0.3 | 0.8 | 4.0 |
| QT4  | 0.8 | 0.8 | 0.8 | 0.8 | 0.6 | 3.8 |
| QT5  | 1.0 | 0.8 | 0.6 | 0.6 | 0.2 | 3.2 |
| QT6  | 0.4 | 0.6 | 0.4 | 0.6 | 0.2 | 2.2 |
| QT7  | 1.0 | 1.0 | 1.0 | 0.4 | 0.8 | 4.0 |
| QT8  | 0.6 | 0.8 | 0.8 | 0.4 | 0.0 | 2.6 |
| QT9  | 1.0 | 0.6 | 0.6 | 0.6 | 0.4 | 3.4 |
| QT10 | 1.0 | 1.0 | 0.9 | 0.7 | 0.4 | 4.0 |
| QT11 | 0.7 | 0.3 | 0.5 | 0.5 | 0.2 | 2.2 |
| QT12 | 0.0 | 0.2 | 0.2 | 0.0 | 0.2 | 0.6 |
| QT13 | 0.2 | 0.2 | 0.4 | 0.0 | 0.0 | 0.8 |

Each score represents the average human rating across five evaluation dimensions. All sub-scores range from 0 to 1, and the total possible score per item is 5. "QT" refers to "Question Type" (e.g., QT1 = Question Type 1).